%% file: CameraReady2027.tex
\documentclass[letterpaper]{article} 
\usepackage{aaai2027}  
\nocopyright
\usepackage[hyphens]{url}  
\usepackage{graphicx} 
\usepackage{natbib}  
\usepackage{caption} 
\usepackage{algorithm}
\usepackage{algorithmic}
\usepackage{placeins}
\usepackage{tabularx}
\usepackage{amsmath}
\usepackage{amssymb}
\usepackage{booktabs}
\usepackage{multirow}
\usepackage{graphicx}
\usepackage{float}
\usepackage{newfloat}
\usepackage{listings}
\DeclareCaptionStyle{ruled}{labelfont=normalfont,labelsep=colon,strut=off} 
\floatstyle{ruled}
\newfloat{listing}{tb}{lst}{}
\floatname{listing}{Listing}

\usepackage{booktabs}

\title{Temporal GRPO: Beyond Trajectory-Level Credit in Vision-Language-Action Reinforcement Learning}
\author {
    Yao Zhou\textsuperscript{\rm 1,\rm 2}\equalcontrib,
    Hang Gao\textsuperscript{\rm 1,\rm 2}\equalcontrib,
    Fengge Wu\textsuperscript{\rm 1,\rm 2},
    Changwen Zheng\textsuperscript{\rm 1,\rm 2}
    Wenwen Qiang\textsuperscript{\rm 1,\rm 2}\corresponding
}
\affiliations {
    \textsuperscript{\rm 1}Institute of Software Chinese Academy of Sciences, Beijing, China\\
    \textsuperscript{\rm 2}University of Chinese Academy of Sciences, Beijing, China\\
}

\begin{document}

\maketitle

\begin{abstract}
Outcome-driven reinforcement learning offers a scalable way to post-train vision-language-action (VLA) policies from sparse task-success feedback. In common GRPO-based VLA post-training, one rollout-level advantage is applied to every action in the trajectory. A rollout that completes several valid stages but fails later can therefore penalize the actions that produced its earlier progress. We call this trajectory-level credit aliasing. Temporal GRPO addresses this problem by constructing detectable task stages, aligning each rollout with stage-specific action intervals, and comparing only rollouts that have entered the same stage. The resulting stage advantages are applied to their corresponding intervals in a single policy update. On RoboTwin 2.0, Temporal GRPO improves task success and sample efficiency, with consistent gains across task horizons. Controlled updates on LIBERO-Long preserve shared prerequisite stages and concentrate improvement at the first stage where rollout outcomes diverge.
\end{abstract}


\section{Introduction}
\input{Chapters/1_Introduction}

\section{Related Works}
\input{Chapters/2_RelatedWorks}

\section{Problem Formulation and Analysis}
\input{Chapters/3_Motivation}

\section{Method}
\input{Chapters/4_Method}

\section{Experiments}
\input{Chapters/5_Experiments}
\section{Conclusion}
\input{Chapters/6_Conclusion}

\bibliography{Chapters/8_references} 
\end{document}

%% file: Chapters/1_Introduction.tex


In recent years, Vision-Language-Action (VLA) models have demonstrated strong potential for multi-task robotic manipulation by jointly modeling visual observations, language instructions, and robot actions, while differences in embodiments, scene distributions, and task objectives still require downstream adaptation~\cite{brohan2023rt2visionlanguageactionmodelstransfer,kim2024openvla,li2024cogactfoundationalvisionlanguageactionmodel,nvidia2025gr00tn1openfoundation}.
To reduce the dependence on additional expert demonstrations, recent work has begun to post-train pre-trained VLAs with reinforcement learning, allowing the policy to improve through environment interaction and task feedback~\cite{li2025simplevla,chen2026lastr1reinforcingroboticmanipulation,lu2025vlarlmasterfulgeneralrobotic,zhang2025grapegeneralizingrobotpolicy,chen2025conrftreinforcedfinetuningmethod,black2026pi0visionlanguageactionflowmodel}.
Representative outcome-driven VLA-RL methods typically sample multiple complete trajectories for the same task, compute group-relative advantages from their final success or failure, and use these advantages to update the sampled action sequences. 


However, a final-failure rollout may have already completed several preceding stages correctly and fail only at a later stage.
For example, early and late failure rollouts can receive the same scalar trajectory advantage despite exhibiting substantially different task progress.
When the negative advantage is broadcast across the executed action sequence, preceding behaviors that produced valid stage progress are suppressed together with the actions responsible for the failure.
We refer to this collapse of different stage outcomes into the same trajectory signal as trajectory-level credit aliasing, which misaligns policy updates with local task progress and weakens sample efficiency and subskill stability in long-horizon manipulation.


This credit aliasing arises because the final outcome of a long-horizon task aggregates multiple stage outcomes, causing rollouts with different task-progress patterns to be mapped to the same failure label and trajectory advantage.
Consequently, directly comparing rollouts that complete different preceding stages and fail at different points yields a trajectory-level advantage that reflects only their overall outcome difference, without identifying the stage from which this difference arises.
We observe that rollouts completing the same prerequisite stages share the same task progress and face the same current-stage completion problem, thereby forming a more comparable local group.
Under this shared condition, rollouts that enter the same stage and exhibit different stage outcomes can provide an informative group-relative ranking signal for that stage.
Accordingly, advantage estimation can be reformulated from unconditional comparison of complete-trajectory outcomes into stage-conditioned comparison under shared prerequisite progress, with each stage advantage assigned to the action interval producing the corresponding stage outcome.

In this paper, we propose Temporal GRPO, a stage-conditioned temporal credit assignment framework for outcome-driven VLA post-training.
Given a task instruction and an initial observation, Temporal GRPO constructs an ordered sequence of detectable task stages and aligns each complete rollout with the corresponding stages and action intervals.
For each stage, the method compares only rollouts that satisfy the shared prerequisites and actually enter that stage, computing group-relative advantages from their stage-completion outcomes.
The resulting stage advantages are assigned to the action intervals producing the corresponding stage outcomes, yielding piecewise temporal credit while preserving the final task-success objective, complete rollouts, and a single VLA policy.
Our main contributions are:
\begin{itemize}
    \item We characterize and formulate trajectory-level credit aliasing in
    outcome-driven VLA reinforcement learning, where rollouts with different
    stage progress can receive the same final-outcome advantage, causing
successful preceding actions and later failed actions to be updated with the
same advantage.

    \item We introduce Temporal GRPO, which constructs detectable ordered stages, groups rollouts by entered stage, and assigns each stage-relative advantage only to the corresponding action interval.

    \item Under matched training budgets, Temporal GRPO achieves a 75.8\% macro-average success rate on RoboTwin 2.0, outperforming the strongest controlled baseline by 7.0 percentage points and delivering consistent gains of 6.2-8.3 points across all task horizons. Controlled LIBERO-Long updates and ablations further show that these gains arise from preserving acquired preceding behaviors and concentrating policy updates on the action intervals responsible for outcome differences.
\end{itemize}

%% file: Chapters/2_RelatedWorks.tex

\subsection{Outcome-Driven Reinforcement Learning for VLA}
\label{sec:rw_vla_rl}


VLA models typically acquire general manipulation capabilities from large-scale robot demonstrations and adapt to specific embodiments and tasks using downstream data; recent studies further employ reinforcement learning post-training to overcome the limited coverage of offline demonstrations through environment interaction and task feedback \cite{kim2024openvla,zhang2025reinbot,li2025simplevla,tan2025interactive}.
Among these approaches, outcome-driven methods directly optimize policies using complete-task success signals and avoid training an additional value model, making them attractive for large-scale VLA post-training.


Representative methods sample multiple complete rollouts under the same task or comparable initial conditions and estimate group-relative advantages from their final outcomes \cite{li2025simplevla,zhang2025reinbot}.
However, these methods primarily distinguish policy performance at the complete-rollout level, causing different stages and actions within a trajectory to share a uniform or highly coupled credit signal.
Our work instead restructures the units of outcome comparison and temporal credit assignment in outcome-driven GRPO without modifying the VLA architecture or introducing a new value model.

\subsection{Temporal Credit Assignment with Structured Task Progress}
\label{sec:rw_structured_credit}


To address sparse or delayed rewards, Hindsight Experience Replay relabels goals achieved in failed trajectories, RUDDER redistributes trajectory returns toward critical timesteps, and related methods recover local contributions through trajectory alignment or learned return models \cite{andrychowicz2017hindsight,arjona2019rudder,patil2020align,zhang2023interpretable}.
Segment-level feedback and stage-aware rewards increase the temporal resolution of supervision, while local group-relative methods estimate micro-level advantages by regrouping similar states or decisions \cite{du2025reinforcement,chen2025sarm,feng2026group}.
Continuous robotic rollouts, however, rarely visit identical visual states and may differ substantially in motion path, stage duration, and execution speed, making raw-state-based cross-rollout grouping unreliable.


Reward Machines represent task progress through high-level events and state transitions, while object- and scene-graph methods abstract continuous robotic states and trajectories using structured relations \cite{icarte2018using,furelos2023hierarchies,sieb2020graph,kumar2023graph,huang2024virl,zhang2023dual}.
These representations are primarily used for task specification, planning, reward construction, or trajectory segmentation, rather than directly supporting advantage estimation in group-relative VLA optimization.
We use verifiable task stages to align action intervals with equivalent task progress across rollouts and compute relative advantages within stage-conditioned groups, thereby reconstructing trajectory-level credit into stage-specific temporal credit.

%% file: Chapters/3_Motivation.tex
In this section, we first formulate outcome-driven VLA post-training under GRPO
and make explicit how a trajectory-level advantage is assigned
across all actions in a rollout. We then analyze how this
assignment maps distinct stage-progress patterns to the same
credit signal, giving rise to trajectory-level credit aliasing
and motivating stage-conditioned temporal credit assignment.

\subsection{Problem Formulation}
We formulate a VLA model as a parameterized policy
$\pi_{\theta}$ that maps the multimodal state $s_t$ at timestep $t$ to a
robot action $a_t$:
\begin{equation}
    a_t \sim \pi_{\theta}(a_t \mid s_t).
    \label{eq:vla_policy}
\end{equation}
Here, $s_t$ denotes the multimodal input available to the policy, including
the current visual observation, the natural-language task instruction, and
optional proprioceptive information.
For VLA policies that predict multiple future control commands at each
decision step, the policy outputs an action chunk of horizon $H$:
\begin{equation}
    \mathbf{a}_{t:t+H-1}
    \sim
    \pi_{\theta}
    \left(
        \mathbf{a}_{t:t+H-1}
        \mid s_t
    \right).
    \label{eq:action_chunk}
\end{equation}
For notational simplicity, we use $a_t$ to denote either a single action or an action chunk.

Reinforcement learning post-training further optimizes the policy through interaction with the target environment.
During environment interaction, the policy generates a trajectory of length
$T$,
\begin{equation}
    \tau
    =
    (s_0,a_0,s_1,a_1,\ldots,s_T),
    \label{eq:trajectory}
\end{equation}
and receives a scalar feedback $r_t$ at each timestep.
The general objective of reinforcement learning post-training is to maximize
the expected discounted return:
\begin{equation}
    \mathcal{J}_{\mathrm{RL}}(\theta)
    =
    \mathbb{E}_{\tau\sim\pi_{\theta}}
    \left[
        \sum_{t=0}^{T-1}
        \gamma^{t} r_t
    \right],
    \label{eq:rl_objective}
\end{equation}
where $\gamma\in[0,1]$ is the discount factor.
In the outcome-driven setting considered in this work, environment feedback is primarily determined by the final task outcome and summarized as a trajectory-level outcome reward $R(\tau)$.


In the outcome-driven VLA reinforcement learning setting considered in this work, Group Relative Policy Optimization (GRPO) samples multiple complete rollouts from the same initial task state and estimates their relative advantages from the outcomes within the rollout group.
Specifically, given an initial state $s_0$ sampled from the task distribution
$\mathcal{D}$, the old policy $\pi_{\theta_{\mathrm{old}}}$ generates a group
of $G$ action sequences:
\begin{equation}
    s_0 \sim \mathcal{D},
    \qquad
    \{a_i\}_{i=1}^{G}
    \sim
    \pi_{\theta_{\mathrm{old}}}(\cdot \mid s_0),
    \label{eq:grpo_sampling}
\end{equation}
The $i$-th rollout is represented as
\begin{equation}
    \tau_i
    =
    \left(
        s_{i,0},
        a_{i,1},
        s_{i,1},
        \ldots,
        a_{i,T_i},
        s_{i,T_i}
    \right),
    \label{eq:rollout_sequence}
\end{equation}
where $T_i$ denotes the number of executed actions in rollout $\tau_i$.

Let $R_i$ denote the trajectory-level outcome reward of the $i$-th rollout.
The group-relative advantage is computed by normalizing the outcome reward
within the rollout group:
\begin{equation}
    \widehat{A}_i
    =
    \frac{
        R_i-\operatorname{mean}
        \left(
            \{R_j\}_{j=1}^{G}
        \right)
    }{
        \operatorname{std}
        \left(
            \{R_j\}_{j=1}^{G}
        \right)
        +\epsilon
    }.
    \label{eq:grpo_group_advantage}
\end{equation}
For the action generated at timestep $t$ in the $i$-th rollout, the policy
probability ratio is defined as
\begin{equation}
    r_{i,t}(\theta)
    =
    \frac{
        \pi_{\theta}
        \left(
            a_{i,t}
            \mid
            s_{i,t}
        \right)
    }{
        \pi_{\theta_{\mathrm{old}}}
        \left(
            a_{i,t}
            \mid
            s_{i,t}
        \right)
    }.
    \label{eq:grpo_policy_ratio}
\end{equation}
The policy is optimized using the following objective:
\begin{equation}
\begin{aligned}
\mathcal{J}_{\mathrm{GRPO}}(\theta)
=&\;
\mathbb{E}_{
\substack{
s_0\sim\mathcal{D},
\{a_i\}_{i=1}^{G}
\sim
\pi_{\theta_{\mathrm{old}}}(\cdot\mid s_0)
}}
\Bigg[
\frac{1}{G}
\sum_{i=1}^{G}
\frac{1}{|a_i|}
\sum_{t=1}^{|a_i|}
\min
\\
&\quad
\Big(
r_{i,t}(\theta)\widehat{A}_i,\,
\operatorname{clip}
\big(
r_{i,t}(\theta),
1-\varepsilon_L,
1+\varepsilon_H
\big)
\widehat{A}_i
\Big)
\Bigg].
\end{aligned}
\label{eq:grpo_objective}
\end{equation}

\subsection{Trajectory-Level Credit Aliasing}
\label{subsec:credit_aliasing}


GRPO provides a value-model-free group-relative optimization scheme in which
the advantage of each rollout is computed from its complete-trajectory
outcome.
Consequently, the same trajectory-level advantage $\widehat{A}_i$ is reused
at every action position:
\begin{equation}
    \widehat{A}_{i,t}
    =
    \widehat{A}_i,
    \qquad
    t=1,\ldots,|a_i|.
    \label{eq:trajectory_advantage_broadcast}
\end{equation}
The complete rollout serves as both the comparison unit for group-relative advantage estimation and the temporal unit over which the advantage is assigned.


Long-horizon robotic tasks commonly contain multiple local stages with
distinct physical semantics, such as reaching the target object, establishing
a stable grasp, transporting the object, and completing placement.
We represent the stage progress of the $i$-th rollout as
\begin{equation}
    \mathbf{z}_i
    =
    \left(
        z_{i,1},
        z_{i,2},
        \ldots,
        z_{i,K}
    \right),
    \qquad
    z_{i,k}\in\{0,1\},
    \label{eq:event_progress}
\end{equation}
where $z_{i,k}=1$ indicates that the $k$-th task stage is achieved.


GRPO computes group-relative advantages from complete-trajectory outcome
rewards, and the stage-progress vector $\mathbf{z}_i$ does not directly enter
the advantage estimation.
Consequently, two rollouts in the same group can satisfy
\begin{equation}
\begin{aligned}
    &i,j\in\{1,\ldots,G\},
    \qquad
    \mathbf{z}_i \neq \mathbf{z}_j,
    \qquad
    R_i = R_j,
    \\
    &\hspace{3.2cm}
    \Longrightarrow
    \widehat{A}_i = \widehat{A}_j.
\end{aligned}
\label{eq:trajectory_credit_aliasing}
\end{equation}
For example, one rollout may complete grasping and transportation and fail
only during final placement, whereas another rollout may fail before
establishing a stable grasp.
Under binary outcome rewards, both rollouts receive the same failure reward,
and their different stage-progress patterns are therefore not reflected in
their group-relative advantages.

When a rollout group contains both successful and failed trajectories,
rollouts with the same failure reward receive the same negative
trajectory-level advantage.
According to Eq.~\eqref{eq:trajectory_advantage_broadcast}, this negative
advantage is assigned to all executed actions in the failed rollout,
including actions that completed preceding stages and actions executed during
the later failed stage.
We refer to the combination of mapping different stage-progress patterns to
the same trajectory-level advantage and uniformly assigning that advantage
to all executed actions as
\emph{trajectory-level credit aliasing}.
This analysis highlights the need for credit estimation in long-horizon
robot reinforcement learning to preserve distinctions in stage progress and
the temporal location of failure.

%% file: Chapters/4_Method.tex

\begin{figure*}[t]
    \centering
    \includegraphics[
        width=0.98\textwidth
    ]{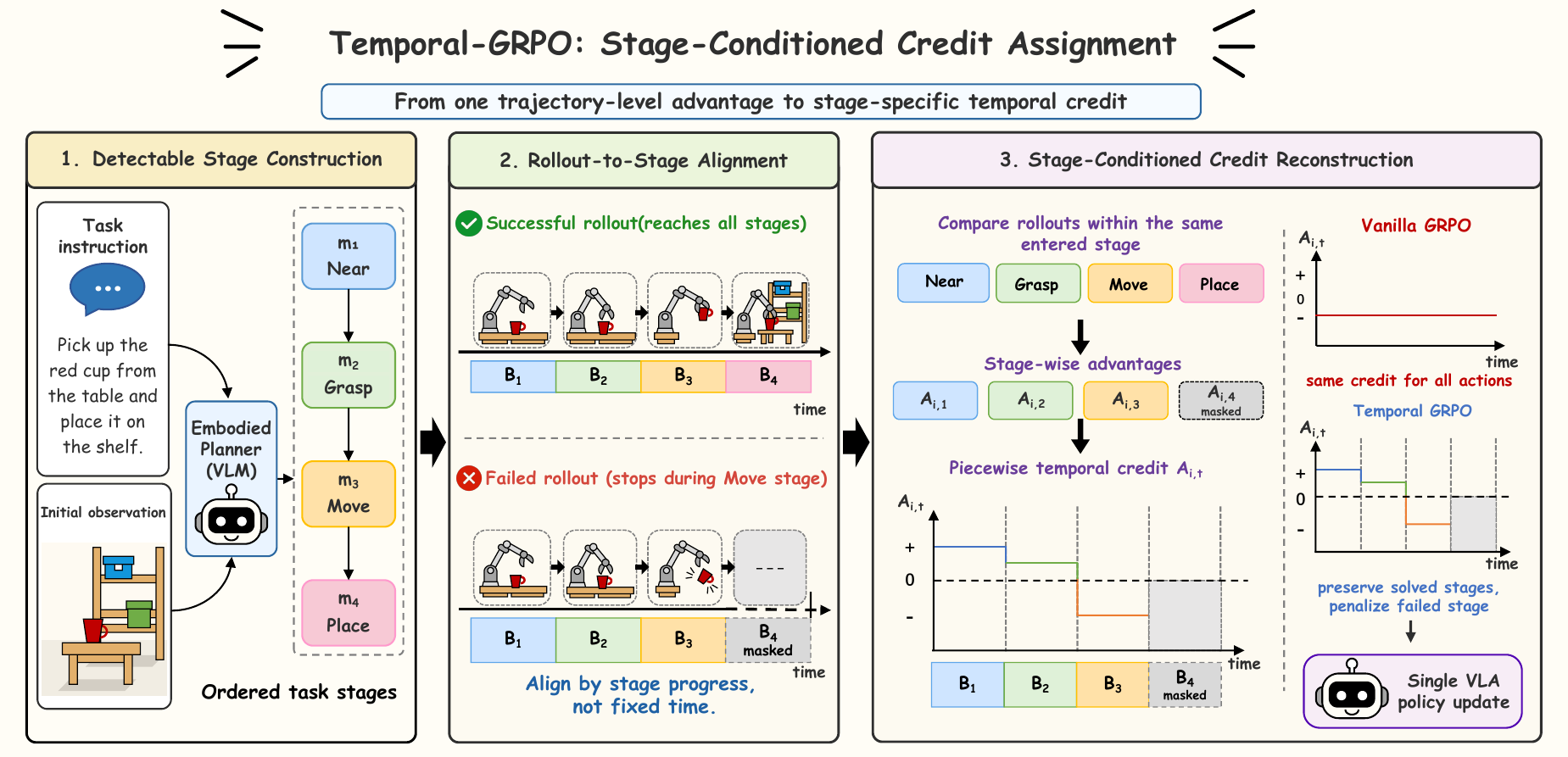}
    \caption{
    Overview of Temporal GRPO, which aligns rollouts with detectable task stages
    and reconstructs stage-specific temporal advantages for VLA policy optimization.
    }
    \label{fig:temporal_grpo_framework}
\end{figure*}
\subsection{Overview}
\label{subsec:method_overview}

To address trajectory-level credit aliasing, we propose Temporal GRPO, which constructs an ordered sequence of detectable task stages from the task
instruction and initial visual observation, and aligns each complete rollout with the corresponding stage-wise action intervals.
For each stage, Temporal GRPO compares only rollouts that have completed their prerequisite stages and entered the current stage, and computes a group-relative advantage from their stage-completion outcomes.
The resulting stage advantage is assigned only to its corresponding action interval, yielding a temporally varying advantage over task progress while optimizing a single VLA policy with a unified GRPO objective.

\subsection{Task-Conditioned Stage Generation}
\label{subsec:stage_generation}


To distinguish different task progress within a complete rollout, we
construct an ordered sequence of task stages from the task instruction $l$
and initial visual observation $o_0$:
\begin{equation}
    M
    =
    F_{\mathrm{stage}}(l,o_0)
    =
    (m_1,m_2,\ldots,m_K),
    \label{eq:credit_stage_generation}
\end{equation}
where $M$ denotes the resulting stage sequence and $m_k$ denotes the $k$-th
stage used for subsequent rollout alignment and credit assignment.
The stage sequence describes the main progression from the initial state to
the final task objective, without decomposing the task into independently
trained sub-policies.
Each stage has explicit task semantics, an ordered position, and a detectable
completion condition, allowing it to serve as the basic unit for local
outcome comparison and temporal credit assignment.


Specifically, we first employ a frozen RynnBrain-4B model~\cite{dang2026rynnbrainopenembodiedfoundation} to propose candidate semantic stages conditioned on the task instruction and initial scene:
\begin{equation}
    \widetilde{M}
    =
    F_{\mathrm{sem}}(l,o_0)
    =
    (\widetilde{m}_1,\widetilde{m}_2,\ldots,
    \widetilde{m}_{\widetilde{K}}),
    \label{eq:semantic_stage_proposal}
\end{equation}
where $\widetilde{M}$ captures a semantic description of the major progress
required to complete the task.
The task instruction specifies the target entities, required operations, and
desired relations, while the initial observation grounds these semantics in
the concrete objects and initial states of the current scene.
The candidate stages provide a semantically meaningful description of task
progress, but their natural-language descriptions may be too abstract or lack
stable detection boundaries and therefore cannot be directly used for
stage-level outcome estimation.

To this end, the Stage Compiler converts the candidate semantic stages into
ordered credit stages that can be used for trajectory detection and credit
assignment:
\begin{equation}
    M
    =
    F_{\mathrm{comp}}
    \left(
        \widetilde{M},l,o_0
    \right)
    =
    (m_1,m_2,\ldots,m_K).
    \label{eq:stage_compilation}
\end{equation}
The compilation process normalizes stage ordering and prerequisite
dependencies, translates abstract stage descriptions into completion
conditions that can be evaluated by the subsequent detection module, and
resolves redundant or ambiguously bounded candidate stages.
The resulting credit stages follow a linear prerequisite structure:
\begin{equation}
    m_1
    \rightarrow
    m_2
    \rightarrow
    \cdots
    \rightarrow
    m_K,
    \label{eq:credit_stage_dependency}
\end{equation}
such that stage $m_k$ is evaluated only after the verified completion of
stage $m_{k-1}$.
The final stage $m_K$ is matched to the original task-success condition,
ensuring that local stage-level credit remains consistent with the complete
task objective.
The output schema, compilation rules, and complete examples of the Stage
Compiler are provided in the supplementary material. For each task
instance, the compiled stage sequence is generated once and shared across all
subsequent rollouts.

\subsection{Rollout-to-Stage Alignment}
\label{subsec:stage_alignment}

The compiled credit stages describe task-level progress, whereas policy
optimization requires identifying the actions associated with each stage in
an actual rollout.
Given the stage sequence $M=(m_1,\ldots,m_K)$ and the $i$-th complete
rollout, we identify where each stage is achieved according to its completion
condition and align the action sequence with the corresponding stage
intervals.
This alignment is determined by task-state transitions. In our simulation experiments, stage predicates are evaluated from privileged simulator states during post-training only; neither the detector nor the privileged states are used at evaluation time.

A relation detector evaluates the completion conditions along the rollout
and begins verifying stage $m_k$ only after stage $m_{k-1}$ has been
confirmed.
For each completed stage, we record the first time at which its completion
condition is stably satisfied:
\begin{equation}
    T_{i,k}
    =
    \min
    \left\{
        t
        \mid
        m_k
        \text{ is stably completed in rollout } i
    \right\}.
    \label{eq:stage_completion_time}
\end{equation}
The stability constraint prevents transient contacts, brief spatial
coincidences, or visual fluctuations from being incorrectly treated as stage
completion.
The relation conditions, stability criteria, and detector implementation are
detailed in the supplementary material.

For a successfully completed stage $m_k$, the corresponding action interval
is defined as
\begin{equation}
    B_{i,k}
    =
    \left(
        T_{i,k-1},
        T_{i,k}
    \right],
    \qquad
    T_{i,0}=0.
    \label{eq:successful_stage_interval}
\end{equation}
This interval contains the actions that advance the rollout from the
completed prerequisite stage $m_{k-1}$ to stage $m_k$.
If rollout $i$ completes $m_{k-1}$ but fails to complete $m_k$ before
termination, the remaining action suffix is assigned to the failed interval
of stage $m_k$:
\begin{equation}
    B_{i,k}
    =
    \left(
        T_{i,k-1},
        T_i
    \right],
    \label{eq:failed_stage_interval}
\end{equation}
where $T_i$ denotes the rollout termination time.
In this case, $m_k$ is treated as the current failed stage, while subsequent stages are considered unreached and are neither assigned valid intervals nor included in later-stage comparisons.
Consequently, each rollout is aligned with a sequence of successful or failed
stage-specific action intervals, which provide the temporal boundaries for
subsequent stage-outcome construction and group-relative advantage
estimation.

\subsection{Stage-Conditioned Advantage Reconstruction and Policy Optimization}
\label{subsec:stage_advantage}


For stage $m_k$, only rollouts that have completed its prerequisite stage
are considered to have entered the current stage and participate in the
corresponding group-relative comparison.
We define the stage participation variable as
\begin{equation}
    V_{i,k}
    =
    \begin{cases}
        1,
        & k=1
        \ \text{or rollout } i \text{ completes } m_{k-1},
        \\
        0,
        & \text{otherwise}.
    \end{cases}
    \label{eq:stage_participation}
\end{equation}
When $V_{i,k}=0$, rollout $i$ has not satisfied the prerequisite of stage
$m_k$ and is therefore excluded rather than treated as a failure at that
stage.


For each rollout satisfying $V_{i,k}=1$, we define the binary stage outcome as
\begin{equation}
    R_{i,k}
    =
    \begin{cases}
        1,
        & \text{rollout } i \text{ completes } m_k,
        \\
        0,
        & \text{rollout } i \text{ fails to complete } m_k.
    \end{cases}
    \label{eq:stage_outcome}
\end{equation}
The outcome of the final stage $m_K$ is identical to the original
task-success signal.
The outcome statistics of stage $m_k$ are computed only over its valid
rollouts:
\begin{equation}
    \mu_k
    =
    \frac{
        \sum_{i=1}^{G}V_{i,k}R_{i,k}
    }{
        \sum_{i=1}^{G}V_{i,k}
    },
    \qquad
    \sigma_k
    =
    \sqrt{
        \frac{
            \sum_{i=1}^{G}
            V_{i,k}
            \left(R_{i,k}-\mu_k\right)^2
        }{
            \sum_{i=1}^{G}V_{i,k}
        }
    }.
    \label{eq:stage_outcome_statistics}
\end{equation}
The stage-conditioned group-relative advantage is then given by
\begin{equation}
    \widehat{A}_{i,k}
    =
    \frac{
        R_{i,k}-\mu_k
    }{
        \sigma_k+\epsilon
    },
    \qquad
    V_{i,k}=1.
    \label{eq:stage_conditioned_advantage}
\end{equation}
If a stage contains no valid rollouts or all of its valid rollouts share the
same outcome, it provides no relative ranking signal and is skipped in the
current policy update.

We assign each stage advantage only to the action interval that produces the
corresponding stage outcome, thereby reconstructing a temporally varying
advantage from the original trajectory-level signal:
\begin{equation}
    \widehat{A}_{i,t}
    =
    \sum_{k=1}^{K}
    \mathbb{I}
    \left[
        t\in B_{i,k}
    \right]
    \widehat{A}_{i,k}.
    \label{eq:temporal_stage_advantage}
\end{equation}
Since the stage intervals are ordered and non-overlapping, each action
inherits at most one valid stage advantage.
For a VLA policy that predicts action chunks, each action chunk and its
action tokens inherit the advantage of the stage interval to which they
belong.


Finally, we replace the uniformly broadcast trajectory-level advantage
$\widehat{A}_i$ with the stage-conditioned temporal advantage
$\widehat{A}_{i,t}$:
\begin{multline}
    \mathcal{J}_{\mathrm{Temporal\mbox{-}GRPO}}(\theta)
    =
    \mathbb{E}
    \Bigg[
    \frac{1}{G}
    \sum_{i=1}^{G}
    \frac{1}{|a_i|}
    \sum_{t=1}^{|a_i|}
    \\
    \min
    \Big(
        r_{i,t}(\theta)\widehat{A}_{i,t},
        \operatorname{clip}
        \big(
            r_{i,t}(\theta),
            1-\varepsilon_L,
            1+\varepsilon_H
        \big)
        \widehat{A}_{i,t}
    \Big)
    \Bigg].
    \label{eq:temporal_grpo_objective1}
\end{multline}
This objective preserves complete-rollout sampling, policy likelihood ratios,
and the clipped GRPO optimization mechanism, while refining credit comparison
and assignment from complete trajectories to stage-specific action intervals.
All stage intervals are jointly optimized within the same training batch and
a unified objective to update a single VLA policy, rather than performing
independent reinforcement learning for different stages.

%% file: Chapters/5_Experiments.tex


We evaluate three questions: whether Temporal GRPO improves final task success, whether it learns more efficiently under a fixed interaction budget, and whether its policy updates are localized to the stage that separates successful and failed rollouts. RoboTwin 2.0 is used for task-level performance and learning curves, while LIBERO-Long is used for controlled credit-assignment analysis and ablations.
\begin{table*}[!t]
    \centering
    \small
    \caption{
        Task success rates (\%) on RoboTwin 2.0 across different task
        horizons. All RL methods are initialized from the same publicly
        released task-specific OpenVLA-OFT SFT checkpoints~\cite{li2025simplevla} and trained
        under identical budgets. RL results are reported as
        mean$\pm$standard deviation over three training seeds.
    }
    \label{tab:robotwin_main_results}

    \setlength{\tabcolsep}{9pt}
    \renewcommand{\arraystretch}{1.02}

    \begin{tabular*}{\textwidth}{
        @{\extracolsep{\fill}}
        l
        c
        c
        c
        c
        @{}
    }
        \toprule
        Method
        & Short Horizon
        & Medium Horizon
        & Long \& Extra-Long
        & Macro Avg. \\
        \midrule

        $\pi_{0}$
        & 45.5
        & 58.8
        & 43.3
        & 49.2 \\

        RDT-1B
        & 24.5
        & 47.8
        & 27.8
        & 33.3 \\

        \midrule

        OpenVLA-OFT (SFT)
        & 21.3
        & 47.1
        & 46.5
        & 38.3 \\

        \midrule

        Trajectory-GRPO
        & $37.8 \pm 1.8$
        & $52.6 \pm 1.6$
        & $48.7 \pm 1.9$
        & $46.4 \pm 1.3$ \\

        TGRPO
        & $43.9 \pm 1.6$
        & $58.4 \pm 1.5$
        & $54.1 \pm 1.7$
        & $52.1 \pm 1.1$ \\

        Stage-Reward GRPO
        & $52.7 \pm 1.4$
        & $64.2 \pm 1.3$
        & $60.8 \pm 1.5$
        & $59.2 \pm 1.0$ \\

        SimpleVLA-RL
        & $\underline{64.9 \pm 1.2}$
        & $\underline{72.5 \pm 1.0}$
        & $\underline{69.0 \pm 1.3}$
        & $\underline{68.8 \pm 0.9}$ \\

        Temporal GRPO
        & $\mathbf{73.2 \pm 0.9}$
        & $\mathbf{79.0 \pm 0.8}$
        & $\mathbf{75.2 \pm 1.1}$
        & $\mathbf{75.8 \pm 0.7}$ \\

        \bottomrule
    \end{tabular*}
\end{table*}
\subsection{Experimental Setup}
\label{subsec:experimental_setup}

\paragraph{Benchmarks.}
We evaluate overall task performance and reinforcement learning sample
efficiency on RoboTwin 2.0~\cite{chen2025robotwin20scalabledata}.
We further conduct controlled credit-assignment analyses and ablation studies
on LIBERO-Long~\cite{liu2023liberobenchmarkingknowledgetransfer} tasks with clear stage dependencies.
RoboTwin 2.0 supports evaluation across task horizons, while LIBERO-Long
enables controlled analysis of shared preceding stages and the first stage
at which rollout outcomes diverge.
Complete results on the four standard LIBERO suites are provided in the supplementary material.

\paragraph{Baselines.}
We compare with the representative VLA baselines $\pi_0$~\cite{black2026pi0visionlanguageactionflowmodel} and RDT-1B~\cite{liu2025rdt1bdiffusionfoundationmodel}, together with the VLA-RL methods SimpleVLA-RL~\cite{li2025simplevla} and TGRPO~\cite{chen2025tgrpofinetuningvisionlanguageactionmodel}.
For the controlled reinforcement learning comparison, SimpleVLA-RL, TGRPO, Trajectory-GRPO, Stage-Reward GRPO, and Temporal GRPO are initialized from the same task-specific OpenVLA-OFT SFT checkpoint.
Trajectory-GRPO assigns the final-outcome advantage to the complete action sequence, while Stage-Reward GRPO converts the same detected stage progress into scalar rewards but retains trajectory-level advantage assignment.
All controlled reinforcement learning methods use the same training scenarios, rollout budget, environment-interaction budget, policy-update budget, and evaluation protocol.
\paragraph{Implementation Details.}
For each RoboTwin 2.0 task, we use the publicly released task-specific OpenVLA-OFT SFT checkpoint as the common warm-start for all controlled reinforcement learning methods.
Trajectory-GRPO, Stage-Reward GRPO, and Temporal GRPO share the same exploration and optimization hyperparameters, isolating the effect of the credit-assignment rule.
SimpleVLA-RL and TGRPO use their recommended configurations under matched rollout, environment-interaction, and policy-update budgets.
For each task specification, a frozen RynnBrain-4B model generates candidate semantic stages once before reinforcement learning, which are reused across all rollouts of that task.
The Stage Compiler converts the candidate stages into credit stages with prerequisite dependencies and detectable completion conditions, after which the relation detector identifies their first completion times and constructs the corresponding action intervals.
Privileged simulator states and stage-related modules are used only for credit construction during reinforcement learning post-training and are not invoked during policy evaluation.
Complete task stages, relation predicates, training hyperparameters, and evaluation configurations are provided in the supplementary material.
\paragraph{Evaluation Metrics.}
The primary metric is task success rate. Controlled reinforcement learning results are reported as the mean and standard deviation over three independent training seeds, with each checkpoint evaluated on a fixed number of held-out rollouts per task.
Sample efficiency is measured by the task success rate under a fixed
environment-interaction budget.
For the controlled credit-assignment analysis, we additionally report changes
in stage-conditional completion probabilities before and after a policy
update.

\begin{figure*}[t]
    \centering

    \begin{minipage}[t]{0.48\textwidth}
        \centering
        \includegraphics[
            width=\linewidth
        ]{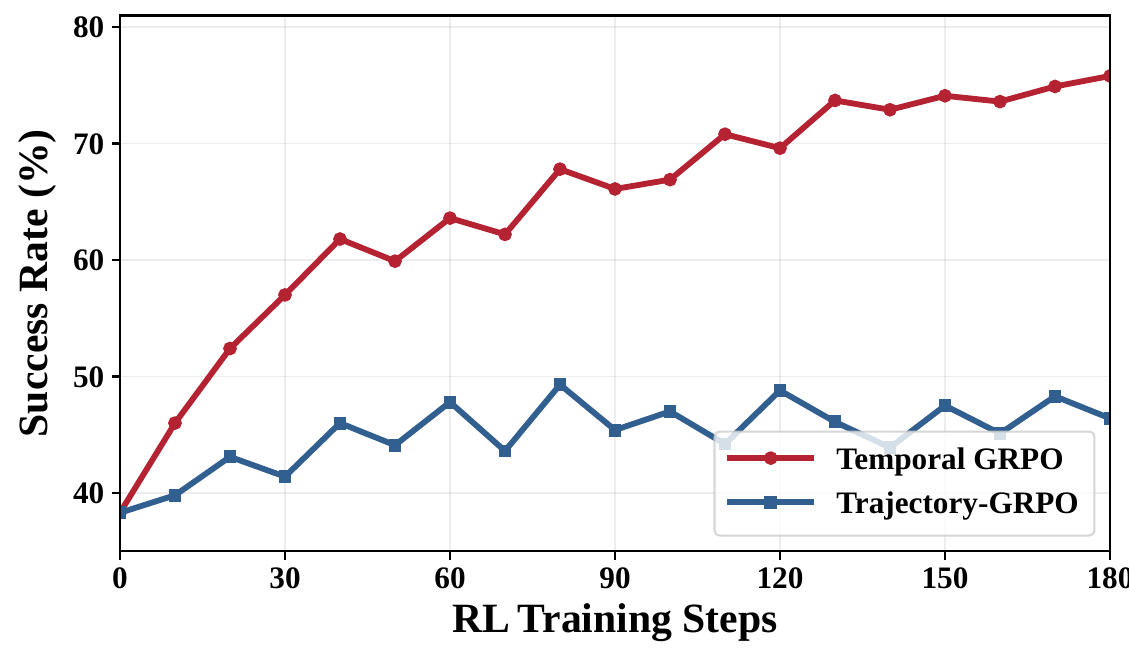}
        
        \vspace{-2mm}
        \centerline{\small (a) All RoboTwin 2.0 tasks}
    \end{minipage}
    \hfill
    \begin{minipage}[t]{0.48\textwidth}
        \centering
        \includegraphics[
            width=\linewidth
        ]{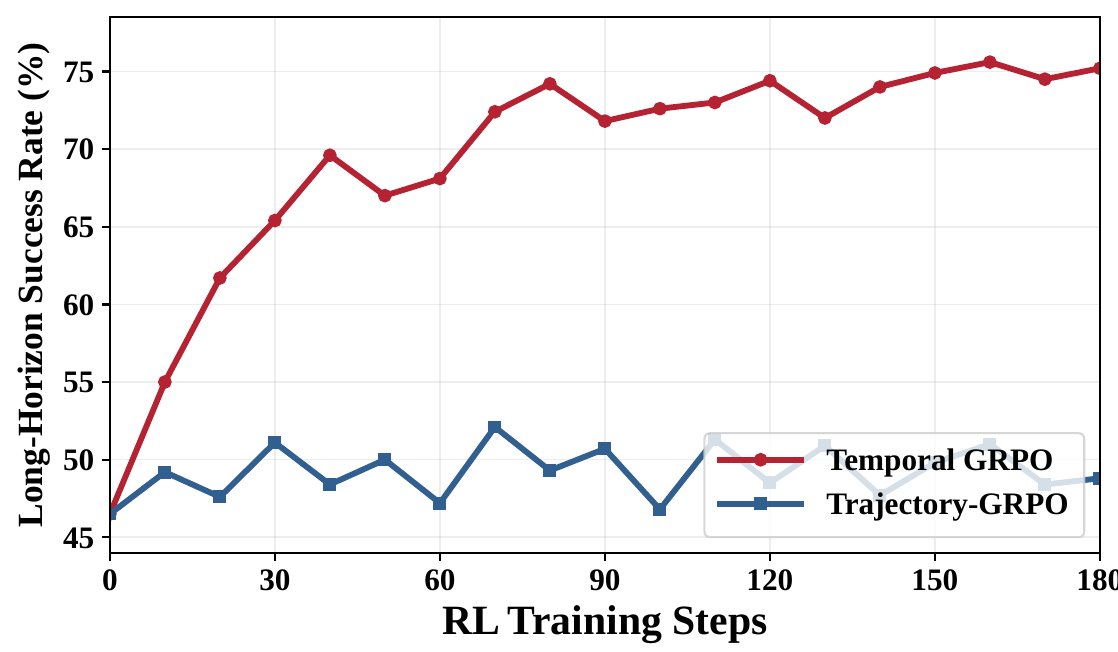}
        
        \vspace{-2mm}
        \centerline{\small (b) Long and Extra-Long tasks}
    \end{minipage}

    \caption{Sample efficiency of Temporal GRPO and Trajectory-GRPO on RoboTwin 2.0. The curves report mean task success rates over three independent training seeds on (a) all tasks and (b) Long and Extra-Long tasks.}
    \label{fig:sample_efficiency}
\end{figure*}

\begin{figure}[t]
    \centering
    \includegraphics[
        width=\columnwidth
    ]{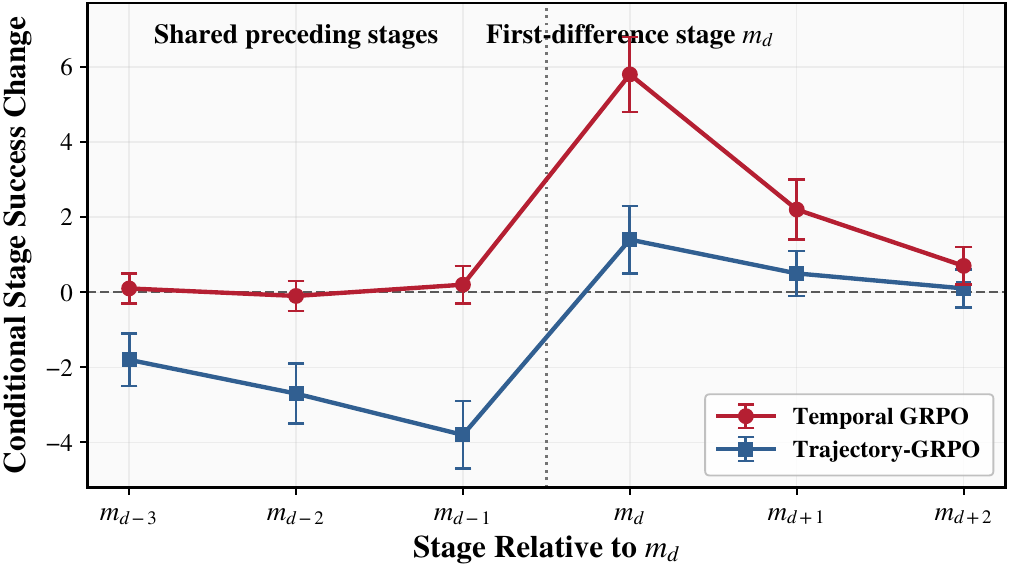}

    \caption{Controlled stage-wise credit assignment on LIBERO-Long.
    Tasks are aligned by the first-difference stage $m_d$. Markers and error bars show the mean $\Delta p_k$ in percentage points
    and one standard deviation}
    \label{fig:controlled_credit}
\end{figure}

\subsection{Overall Performance and Sample Efficiency}
\label{subsec:overall_performance}

Table~\ref{tab:robotwin_main_results} reports task success rates across different RoboTwin 2.0 task horizons. Results for the reinforcement learning methods are presented as the mean and standard deviation over three independent training seeds. Temporal GRPO achieves the highest success rate in every task-horizon group. Its macro-average success rate reaches $75.8\pm0.7$, the best result among all compared methods. The improvement on Long and Extra-Long tasks supports the use of stage-conditioned credit assignment for long-horizon manipulation with sequential stage dependencies.

Figure~\ref{fig:sample_efficiency} shows task success rates as a function of reinforcement learning training steps on all RoboTwin 2.0 tasks and on the Long and Extra-Long subset. All methods use the same number of rollouts per training step, so comparisons at the same training step correspond to matched environment-interaction budgets. Temporal GRPO maintains higher task success rates than Trajectory-GRPO throughout post-training. The efficiency advantage remains clear on the Long and Extra-Long subset, where delayed outcomes make temporal credit assignment more challenging. These results indicate that stage-conditioned credit assignment extracts more effective training signals from the same rollout budget, yielding faster improvement and higher final task success.

\subsection{Controlled Credit Assignment}
\label{subsec:controlled_credit}


Overall task success cannot reveal which task stages are improved or degraded
by a policy update.
To directly examine where credit is assigned, we select matched groups from naturally sampled rollouts. Within each group, trajectories successfully complete the same prerequisite stages and first diverge in their stage-completion outcomes at stage $m_d$.
For stage $m_k$, we measure the update effect using the change in completion
probability among rollouts that enter the stage:
\begin{equation}
\begin{aligned}
\Delta p_k
=
100\Bigg[
&\Pr_{\tau\sim\pi_{\theta^{+}}}
\left(
R_k(\tau)=1
\mid
V_k(\tau)=1
\right)
\\
&-
\Pr_{\tau\sim\pi_{\theta}}
\left(
R_k(\tau)=1
\mid
V_k(\tau)=1
\right)
\Bigg].
\end{aligned}
\label{eq:stage_success_change}
\end{equation}
Here, $V_k(\tau)$ and $R_k(\tau)$ indicate whether rollout $\tau$ enters and completes stage $m_k$, respectively.
Positive and negative values of $\Delta p_k$ indicate improved and degraded
stage completion, measured in percentage points.


Figure~\ref{fig:controlled_credit} aligns different LIBERO-Long tasks by the
first-difference stage $m_d$ and compares $\Delta p_k$ across relative stage
positions.
Trajectory-GRPO produces clear negative changes on the shared preceding
stages, showing that its negative trajectory-level advantage propagates to
already successful action intervals.
Temporal GRPO keeps changes on the preceding stages close to zero and produces
the largest positive improvement at $m_d$, indicating that it preserves
acquired preceding behaviors and concentrates the update on the stage
responsible for the rollout difference.

\subsection{Ablation Studies}
\label{subsec:ablation}


We evaluate all ablation variants on LIBERO-Long using the same policy
initialization and training budget, and report the mean success rate and
standard deviation over three independent seeds.
\begin{table}[t]
    \centering
    \small
    \caption{
        Ablation results on LIBERO-Long under the same training budget.
        Results are reported as mean $\pm$ standard deviation over three seeds.
    }
    \label{tab:ablation}
    \setlength{\tabcolsep}{8pt}
    \renewcommand{\arraystretch}{1.05}
    \begin{tabular}{lc}
        \toprule
        Variant & Success Rate (\%) $\uparrow$ \\
        \midrule
        Temporal GRPO
        & $\mathbf{99.1 \pm 0.4}$ \\
        w/o Stage Compiler
        & $96.8 \pm 0.7$ \\
        Stage-Reward GRPO
        & $94.7 \pm 0.9$ \\
        w/o entered-stage gating
        & $92.5 \pm 1.1$ \\
        w/o same-stage grouping
        & $90.6 \pm 1.3$ \\
        \midrule
        Trajectory-GRPO
        & $88.4 \pm 1.5$ \\
        \bottomrule
    \end{tabular}
\end{table}
Without entered-stage gating, rollouts that have not entered the current stage
are treated as stage failures, resulting in a clear performance degradation.
Removing same-stage grouping causes the largest component-level degradation,
showing that action intervals from different stages do not form a reliable
shared relative comparison group.
The degradation without the Stage Compiler indicates that ordered
prerequisites and detectable completion conditions improve the reliability of
rollout-to-stage alignment.


Stage-Reward GRPO consistently outperforms Trajectory-GRPO, confirming that
stage progress provides useful additional supervision.
Temporal GRPO further achieves the best performance, demonstrating that
stage-conditioned comparison and advantage assignment to the corresponding
action intervals remain essential to the full improvement.

%% file: Chapters/6_Conclusion.tex
This paper introduces Temporal GRPO to address trajectory-level credit
aliasing in outcome-driven VLA post-training.
It constructs detectable ordered task stages, compares rollouts entering the
same stage under shared prerequisite progress, and assigns stage-relative
advantages to the corresponding action intervals.
Experiments on RoboTwin 2.0 demonstrate improved task success and sample efficiency across task horizons, while controlled LIBERO-Long analyses and ablations support the proposed stage-localized credit mechanism.

Temporal GRPO currently relies on reliable stage predicates and a predefined linear stage order, which may limit stage alignment and credit assignment when stage boundaries are ambiguous or when task progress involves branching, repeated stages, or recovery through earlier stages. Future work will explore uncertainty-aware stage detection and dynamic stage graphs to extend temporal credit assignment to more complex manipulation processes.